\documentclass[letterpaper, 10 pt, conference]{ieeeconf}  

\IEEEoverridecommandlockouts                              

\usepackage{graphicx}

\title{\LARGE \bf
Swarm Performance Indicators: Metrics for Robustness, Fault Tolerance, Scalability and Adaptability}

\author{Emma Milner$^{1,2}$, Mahesh Sooriyabandara$^{2}$ and Sabine Hauert$^{1}$
\thanks{$^{1}$University of Bristol, United Kingdom
        {\tt\small emma.milner@bristol.ac.uk}}%
\thanks{$^{2}$Toshiba Research Europe Ltd, 
        Bristol, BS1 4ND
}}

\begin{document}

\maketitle
\thispagestyle{empty}
\pagestyle{empty}
\begin{abstract}
Swarms have distributed control and so are assumed to inherently have superior robustness, scalability and adaptability compared to centralised multi-agent systems. However, these features have generally only been defined qualitatively and there is a lack of quantitative metrics and experimental measures for the claimed parameters. Swarm Performance Indicators are defined here as Key Performance Indicators for swarm features but can be applied to multi-agent systems with centralised control as well. These swarm features are \textit{Robustness, Fault Tolerance, Adaptability }and \textit{Scalability}. Swarm Performance Indicators can be used to highlight the benefits of swarms beyond solely considering task-based performance metrics (e.g. time taken). 
\end{abstract}
\section{Introduction}
Swarm robotics are a form of multi-agent systems that use distributed control. This means that their control is entirely based on local sensing and communication, without global information or centralised control. This gives redundancy \cite{winfield2006}, scaling without increasing complexity \cite{heinrich2022} and the ability to self-organise \cite{schranz2020}. These features of swarm behaviour are often extrapolated by researchers to conclude that their swarm is robust to failures, has good scalability and good adaptability to environmental changes, without giving tests or evidence for these traits. They are given to be inherent in swarm robotics, which is wrong \cite{dorigo2021}. Winfield and Bjerknes (2013) said that this lack of empirical evidence for such claims is dangerous and will need to be addressed if swarms are to move into real-world use. These features will need to be testable and measurable, like Key Performance Indicators (KPIs) that are used in industry. Using currently available KPIs (such as time taken to collect a given number of items in a warehouse) would miss the key features that make a swarm beneficial over a centralised system, such as robustness, scalability and adaptability. Solutions with such features are useful because (e.g.) they mean the swarm is usable out-of-the-box in varying environments. Developing KPIs for these features will avoid swarms being undersold and misunderstood in their transition to industry. In addition, the limits and circumstances in which these traits are present will need to be clearly defined by each swarm researcher when discussing their system. For example, if a researcher says that their swarm is robust to failures, they should say what failure modes the swarm is robust to and how many agents can fail by this mode before the system is no longer considered robust \cite{winfield_bjerknes}. The metrics that are developed in this work are demonstrated on a swarm logistics task to exemplify how they might be used in the real world. These SPIs can also be used by swarm engineers when designing their controllers to increase e.g. robustness. For example, in machine learning techniques for evolving better swarm algorithms, swarm engineers could include the SPIs in fitness functions towards desired robust, scalable and adaptable behaviours. This is not easily done without quantitative metrics. 
\section{Literature review} 
\label{sec:lit_review}
Three traits are explored in this literature review to find a unifying term for each one, based on the state of the art literature on the topic. Many discussions of these features use interchangeable terms and definitions. There are no unifying quantitative definitions for these traits but there are common themes when describing an effect or justifying why a swarm has that trait. Brambilla et al. (2013) agreed that is a ``lack of well-defined metrics" which are needed for verification and validation in swarm robotics. In a recent paper, Dorigo et al. (2021) stated that these properties of swarms are still being taken for granted as inherent in swarms, rather than by given evidence. Winfield and Nembrini (2006) did a full experiment to show how it is unsafe to claim robustness to failures without evidence or explanation of the circumstances where a swarm is robust or not robust. \\
\textbf{State-of-the-art metrics} Hecker and Moses (2015) wrote metrics for error tolerance, flexibility and scalability based on performance (which they termed efficacy, E). Their error tolerance measures the performance change between a strategy developed in the presence of errors and one not. Similarly, their flexibility is the ratio between two strategies applied to resource distribution A when one has been developed with A and the other is developed with resource distribution B. Finally, they evaluated scalability by how much the performance per robot is improved in greater swarm sizes compared to a single robot.\\
\\
\textit{\textbf{Trait (1)} The ability to cope with agent failures}\\
\textbf{Terminology:} This first property is termed by the literature as either \textit{Robustness} \cite{brambilla}, \cite{harwellandgini2020}, \cite{sahin05}, \cite{2schranz2020} or \textit{Fault/Error Tolerance} \cite{winfield_bjerknes}, \cite{dorigo2021}, \cite{heckerandmoses15}, \cite{heinrich2022} (in some sources, these terms are used interchangeably). With the exception of Dorigo et al. (2021) and {\c{S}}ahin (2005) who both include disturbances in the environment in their definition of ``robustness", as opposed to agent failures. However Harwell and Gini (2020) directly contradict this inclusion of environmental factors in the term ``robustness'' by stating it is ``in response to internal, as opposed to environmental stimuli''. \\
\textbf{How to measure Robustness and Fault Tolerance} Firstly, Winfield and Nembrini (2006) found through their study of robustness to failures swarm agents, that it was important to test for several, separate failure modes. Some failure modes may be disastrous if only one agents fail in this way and other failure modes may have barely any effect even when many agents fail. Secondly, there are two factors that are discussed in the literature to do with a swarms' \textit{ability to cope with agent failures}. 
\begin{enumerate}
    \item \textbf{Effect of failed agents on group performance} When failures occur in the swarm, many sources define the ability to cope with failures by the degree of performance lost. Harwell and Gini (2020) state that ``performance should remain the same", {\c{S}}ahin (2005) only requires that the swarm ``operate, although at a lower performance" and Dorigo et al. (2021) allows for a ``graceful degradation of performance in the presence of system faults". This final definition from Dorigo et al. (2021) is used for the definition of the Swarm Performance Indicator for \textit{Robustness}. 
    \item \textbf{Effect of failed agents on working agents} Winfield and Bjerknes (2013) describe self-repair time in their discussion of fault tolerance. Self-repair time is the time taken by working agents in the swarm to remove themselves from the negative influences of the failed agents, known as anchors. The performance of the swarm is degraded further than just the loss of the faulty agent when failed agents ``anchor'' working agents. This is an example of a behaviour caused by failed agents which should be measured and is used for the definition of the Swarm Performance Indicator for \textit{Fault Tolerance}. 
\end{enumerate}
To include the two factors that affect the ability to cope with agent failures, two terms are here defined to be developed into Swarm Performance Indicators for trait \textbf{(1)}:\\
1. If \textit{m} agents out of $N$ total have failed by Failure Mode X, then good \textbf{Robustness} occurs when the percentage change in performance is less than the percentage change in agents (\textit{m/N}).\\
2. If \textit{m} agents out of $N$ total have failed by Failure Mode X, then good \textbf{Fault Tolerance} occurs when the percentage change in performance change due to faults is less than the percentage change in performance of a working swarm with $N-m$ agents.\\
\\
\textit{\textbf{Trait (2)} The ability to cope with changing swarm size}\\
\textbf{Terminology:} This trait is termed across the literature as \textit{Scalability}
\cite{brambilla}, \cite{dorigo2021}, \cite{hamann2022}, \cite{harwellandgini2020}, \cite{heckerandmoses15}, \cite{heinrich2022}, \cite{sahin05}, \cite{2schranz2020}.\\
\textbf{How to measure Scalability:} There are two measures of good scalability used in swarm research in general.
\begin{enumerate}
    \item \textbf{Performance does not decrease with increased size} Dorigo et al. (2021) and {\c{S}}ahin (2005) require performance to remain the same under differing group sizes, for a system to be scalable. Hamann and Reina (2022) has a higher requirement for scalability, ``a scalable system has increasing performance with increasing swarm size".
    \item \textbf{Performance increases \textit{superlinearly} with increased swarm size} Hamann and Reina (2022) and other sources \cite{harwellandgini2020}, \cite{heinrich2022}, continue their discussion of sclability to include (but not require) a superlinear increase in performance for additional agents. If $P_N$ is the initial performance with $N$ agents, then for superlinear scalability $P_{N+1}>P_N+P_N/N$. 
\end{enumerate}
The Swarm Performance Indicator definition for scalability uses the first condition (that performance at least remain the same for $N$ agents as $N+m$) as its minimum condition for scalability. Superlinear scalability is included in the definition as increased scalability:\\
3. A system is \textbf{Scalable} when the performance change is positive with increasing swarm size. The system is \textbf{Superlinearly Scalable} when the performance change per agent is superlinear. \\
\\
\textit{\textbf{Trait (3)} The ability to cope with changing external factors (environmental or task parameters)} \\
\textbf{Terminology:} The definition given in \textbf{(3)} is specific to parameters that can change in the swarm environment or task. It is \textbf{not} meant to describe how well a swarm can be applied to different environments or tasks. The terms that are used to describe this include \textit{Flexibility} \cite{brambilla}, \cite{heckerandmoses15} and \textit{Adaptability} \cite{dorigo2021}, \cite{harwellandgini2020}, \cite{prehofer}, \cite{2schranz2020}. ``Flexibility'' is often used to describe swarms that can be applied across different tasks \cite{dorigo2021}, \cite{sahin05} which is not what trait \textbf{(3)} intends to include. Harwell and Gini (2020) define ``adaptability'' as the ``ability to minimise performance losses under adverse conditions and proportionally exploit beneficial deviations from ideal conditions". Dorigo et al. (2021) define ``adaptability'' (and ``adaptivity'') by the ability to change behaviour to new operating conditions such as obstacles or changing atmospheric conditions. Harwell and Gini (2020) and Dorigo et al. (2021) give definitions which most closely match the intent of \textbf{(3)} and both ``adaptability'' therefore this is the term chosen here.\\
\textbf{How to measure Adaptability:} Harwell and Gini (2020) define an adaptable system as one where performance is gained in changing conditions and minimises any performance lost.\\
4. A system is \textbf{Adaptable} to changes in external parameter, $x$, if performance is gained at all or performance loss is proportional to the change in parameter $x$. Here, $x$ can be an environmental factor (e.g. number of obstacles, brightness of light) or a task parameter (e.g. size of warehouse space, speed of agents). 
\section{Swarm Performance Indicators}
These Swarm Performance Indicators are metrics to describe traits seen in swarm behaviours, with their terminology reasoned using current literature on the topic from Section \ref{sec:lit_review}. The metrics defined here are intended to improve the measurability of swarm behaviours. When we want to know how a system is fairing under faults or changes to the external parameters or changing swarm size, what the user is asking is how are these factors affecting the performance of the system. Given the difference in performance from one condition to another (e.g. increase of 10 s) does not provide enough context. First, it is important to know how much the performance has changed in proportion to the usual performance of the system. This is most easily described by percentage change in performance. For example, `there has been a 10\% loss in performance due to faults in the system' is more informative at a glance than `the performance has gone from 100s to 90s when faults are present'. Secondly, how much the system has changed to cause this performance change is an important factor in determining the effect of the change on the system. If we know that the system has lost 10\% of its original performance due to faults then it is necessary to know more details about the magnitude of these faults. For example, if the system lost 10\% of performance when 1\% of agents failed then this is a worse result than if the same performance loss occurred when 80\% of agents have failed. The 10\% looks bad when the agent loss percentage is low and it looks good when the agent loss percentage is high. \\
\\
\textit{\textbf{Scalability metric}}\\
The swarm trait that the Scalability metric, S, describes is: \textit{The ability to cope with changing swarm size}. This was defined following a literature review (Section \ref{sec:lit_review}) as: A system is \textit{scalable} when the performance change is positive with increasing swarm size. The system is \textit{superlinearly scalable} when the performance change per agent is superlinear. \\
To measure scalability as intended here, the performance should be measured for a given system with $N$ agents and then again with $N+m$ agents. The conditions and all other parameters between the two swarm size tests should be identical. The density of agents in the space will change between the two systems and the size of the space should not be adjusted to keep density constant. Coping with differing densities is a contributing factor in the scalability result. 
When quoting the scalability metric result it is important to also state the conditions which this result was found such as the size of the swarm space and the radius of the agents. \\
\\
\textit{Definition for Scalability}: In the literature, scalability is defined as when no performance is lost by increasing swarm size \cite{dorigo2021}, \cite{sahin05}, which is Condition 1: $P_{N+m} \geq P_N$ and as being good (superlinear) when the performance is increased by more than the performance per agent \cite{harwellandgini2020}, \cite{heinrich2022}. This is, Condition 2: $P_{N+m} > P_N + \frac{m*P_N}{N}$\\
\\
\textit{Derivation of Scalability metric, S}: In a scalable system that moves from $N$ agents to $N+m$ agents, the performance at the two swarm sizes follows Condition 1: $P_{N+m} > P_N$. In a superlinear scalable system going from $N$ to $N+m$ agents, the performance of the larger system satisfies Condition 2: $P_{N+m} > P_N + m*(P_N/N)$. Therefore, Scalability, S has the following derivation where the original swarm performance $P_N$ with $N$ agents is changed to $P_{N+m}$ by the addition of \textit{m} agents. 
\begin{equation}
    \frac{\frac{P_{N+m} - P_N}{P_N}}{\frac{m}{N}}> 1
\end{equation}
\begin{equation}
    S = \frac{\frac{P_{N+m} - P_N}{P_N}}{\frac{m}{N}}
\end{equation}
Such that a scalable system is \textbf{S $>$ 0} which satisfies Condition 1 and a superlinear scalable system satisfies Condition 2 when \textbf{S $>$ 1}. Simplified, this is, 
\begin{equation}
\%\Delta P = \frac{P_{N+m}-P_N}{P_N}
    \label{eq:SdP}
\end{equation}
\begin{equation}
\%\Delta N = \frac{m}{N}
    \label{eq:SdN}
\end{equation}
\begin{equation}
S = \frac{\%\Delta P}{\%\Delta N} 
    \label{eq:S}
\end{equation}
\\
\textit{\textbf{Fault tolerance and robustness metrics}}\\
The swarm trait that these metrics are describing is: \textit{The ability to cope with faulty agents in the system}. The effect of faults on the system has been described in multiple ways within in the literature (see Section \ref{sec:lit_review}). To include the different elements of the effect of faults, two different metrics are proposed: Fault Tolerance and Robustness.
Only one failure mode should be tested at a time so that the metric is clearly specific to that set of parameters, failure mode and number of failures. This supports research into fault tolerance metrics by Winfield and Bjerknes (2013). \\
\textbf{Fault tolerance metric} It is important to examine if the faulty agents are having a negative effect on their neighbouring (working) agents or if their effect is still positive, despite their faults. 
For example, if their wheels have failed, causing them to be static obstacles, they may still be able to communicate sensory information which may help the task more than if they were just removed from the system when they failed. This is examined by this metric, which compares the performance of the equivalent, scaled down swarm performance $P_S$ to the failed swarm performance $P_F$. For $P_S$, $m$ agents have been removed from the swarm space. In the failed swarm, $m$ agents have failed and remain in the swarm space. If the failed swarm performs better than the scaled down swarm then there is an indication that the swarm is tolerant to the fault X with $m$ failed agents because the failed agents still contribute to the performance more than if they were removed altogether.\\
\\
\textit{Definition of Fault Tolerance}: The definition for Fault Tolerance that was reasoned in Section \ref{sec:lit_review}, is: If \textit{m} agents out of $N$ total have failed by Failure Mode X, then good \textit{Fault Tolerance} occurs when the performance change under faults ($\%\Delta P_F$) is better than the performance ($\%\Delta P_S$) of a working swarm with \textit{m} fewer agents than $N$. Therefore, in a fault tolerant system, $\%\Delta P_F > \%\Delta P_S$\\
\\
\textit{Derivation of Fault Tolerance metric, FT}: A fault tolerant system is $\%\Delta P_F > \%\Delta P_S$, where $\%\Delta P_S$ and $\%\Delta P_F$ are
\begin{equation}
    \%\Delta P_S = \frac{P_S - P_o}{P_o}
    \label{eq:dS}
\end{equation}
\begin{equation}
    \%\Delta P_F = \frac{P_F - P_o}{P_o}
    \label{eq:dPF}
\end{equation}
where $P_o$ is the original performance with $N$ agents and no faults. The condition for good Fault Tolerance can be rearranged to $\%\Delta P_F - \%\Delta P_S > 0$. Therefore Fault Tolerance, FT, is: 
\begin{equation}
    FT = \%\Delta P_F - \%\Delta P_S
    \label{eq:FT}
\end{equation}
Where \textbf{FT $>$ 0} is a fault tolerant system.\\
\\
\textbf{Robustness metric} For a system to be counted as robust it is permissible for some performance to be lost when some agents are faulty \cite{sahin05}, \cite{dorigo2021}. How much loss of performance is acceptable has not been defined \cite{winfield_bjerknes} but a reasonable suggestion is given here as a metric for robustness. Dorigo et al. (2021) describes a ``graceful degradation" in performance due to faults. Therefore, in this metric, we determine that a system is robust to fault X if that faulty agent only removes their own performance contribution.\\
\\
\textit{Definition of Robustness}: The definition for Robustness that was reasoned in Section \ref{sec:lit_review} is: If \textit{m} agents out of $N$ have failed by Failure Mode X, then good \textit{Robustness} occurs when the percentage change in performance ($\%\Delta P$) is less than the percentage change in agent number ($\%\Delta N$). A system is robust if the performance $P_F$ with \textit{m} faulty agents out of N total is, $P_F > P_o - \frac{m*P_o}{N}$ where $P_o$ is the performance without faults. Otherwise, if the faulty agents cause more performance loss than agent loss then the system is not robust. \\
\\
\textit{Derivation of Robustness metric, R}: Following the definition given for robustness, the metric R can be derived as follows $\frac{P_F - P_o}{P_o} + \frac{m}{N} > 0$. Which can be simplified to, $\%\Delta P + \%\Delta N > 0$. Therefore, 
\begin{equation}
    R = \%\Delta P + \%\Delta N
    \label{eq:R}
\end{equation}
Where \textbf{R $>$ 0} is robust and $\%\Delta P$ and $\%\Delta N$ (using the original performance $P_o$ with no agent failures and $N$ agents) are 
\begin{equation}
    \%\Delta P = \frac{P_F - P_o}{P_o}
    \label{eq:RdP}
\end{equation}
\begin{equation}
    \%\Delta N = \frac{m}{N}
    \label{eq:RdN}
\end{equation}
%
\\
\textit{\textbf{Adaptability metric}}\\
The swarm trait which this metric describes is: \textit{The ability to cope with changing external parameters.} This was defined following a literature review (Section \ref{sec:lit_review}) as: A system is \textbf{adaptable} to changes in external parameter, $x$, if any performance lost $\%\Delta P$ is less than the proportional change in parameter $x$,  $\%\Delta x$. To measure this metric how it is intended, it is necessary to test the following two cases of the given system. The two cases should have identical parameters and conditions but with one test parameter of interest having a different value between the two. This parameter could be either environmental (e.g. number of obstacles, level of brightness) or task based (e.g. number of targets to find, task time limit). \\
\\
\textit{Definition of Adaptability}: A system is adaptable to changes in parameter, $x$,  if the performance improves from the original performance $P_o$. This is Condition 1, $P_{new} > P_{o}$. If this is not true and performance is lost when moving from $x_o$ to $x_{new}$, then the system is only adaptable if the percentage loss in performance is proportional to the modulus of the percentage change in parameter $x$. The modulus is used because it is irrelevant if parameter $x$ increases or decreases. This is Condition 2, $P_{new} > P_o - P_o*|\frac{x_{new}-x_o}{x_o}|$\\
\\
\textit{Derivation of Adaptability metric, A}: A system is adaptable when Condition 2 is satisfied which can be rearranged to:
\begin{equation}
    \frac{\%\Delta P}{|\%\Delta x|} + 1 > 0 
\end{equation}
\begin{equation}
    A = \frac{\%\Delta P}{|\%\Delta x|} + 1
    \label{eq:A}
\end{equation}
The system is adaptable if \textbf{A $>$ 0} because Condition 2 is satisfied. The adaptability is very good when \textbf{A $>$ 1} as Condition 1 is satisfied, meaning that performance has been gained. $\%\Delta P$ and $\%\Delta x$ are,
\begin{equation}
  \%\Delta P = \frac{P_{new} - P_o}{P_o}
  \label{eq:AdP}
\end{equation}
\begin{equation}
    \%\Delta x = \frac{x_{new}-x_o}{x_o}
    \label{eq:Adx}
\end{equation}
\section{Use case example: Logistics scenario}
\begin{figure}[h]
   \centering
\includegraphics[scale=0.150]{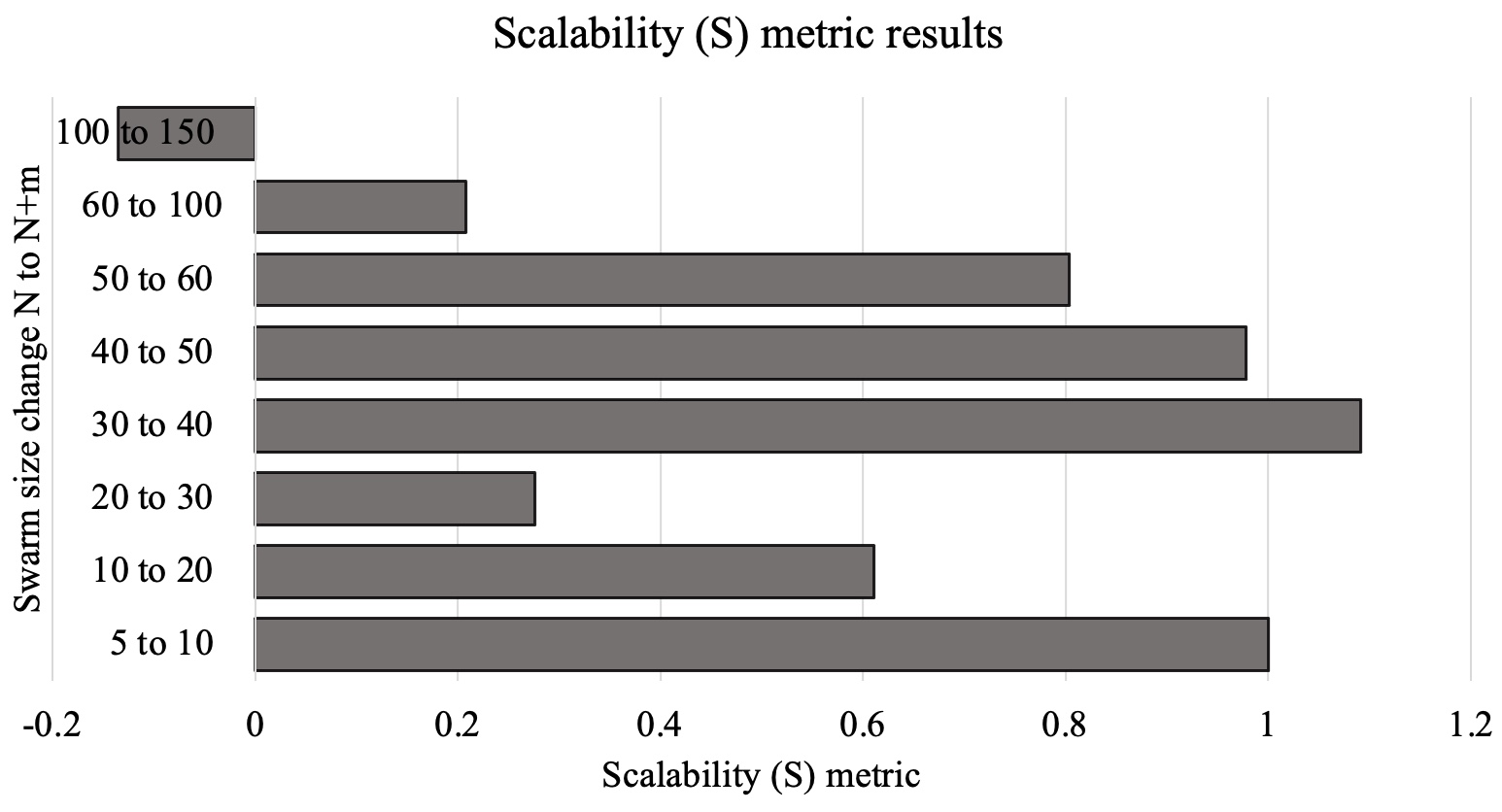}
\caption{Scalability, $S$, for incremental increases in swarm size ($N$ to $N+m$).}
\label{fig:S}
\end{figure}

\begin{figure}[h]
   \centering
\includegraphics[scale=0.11]{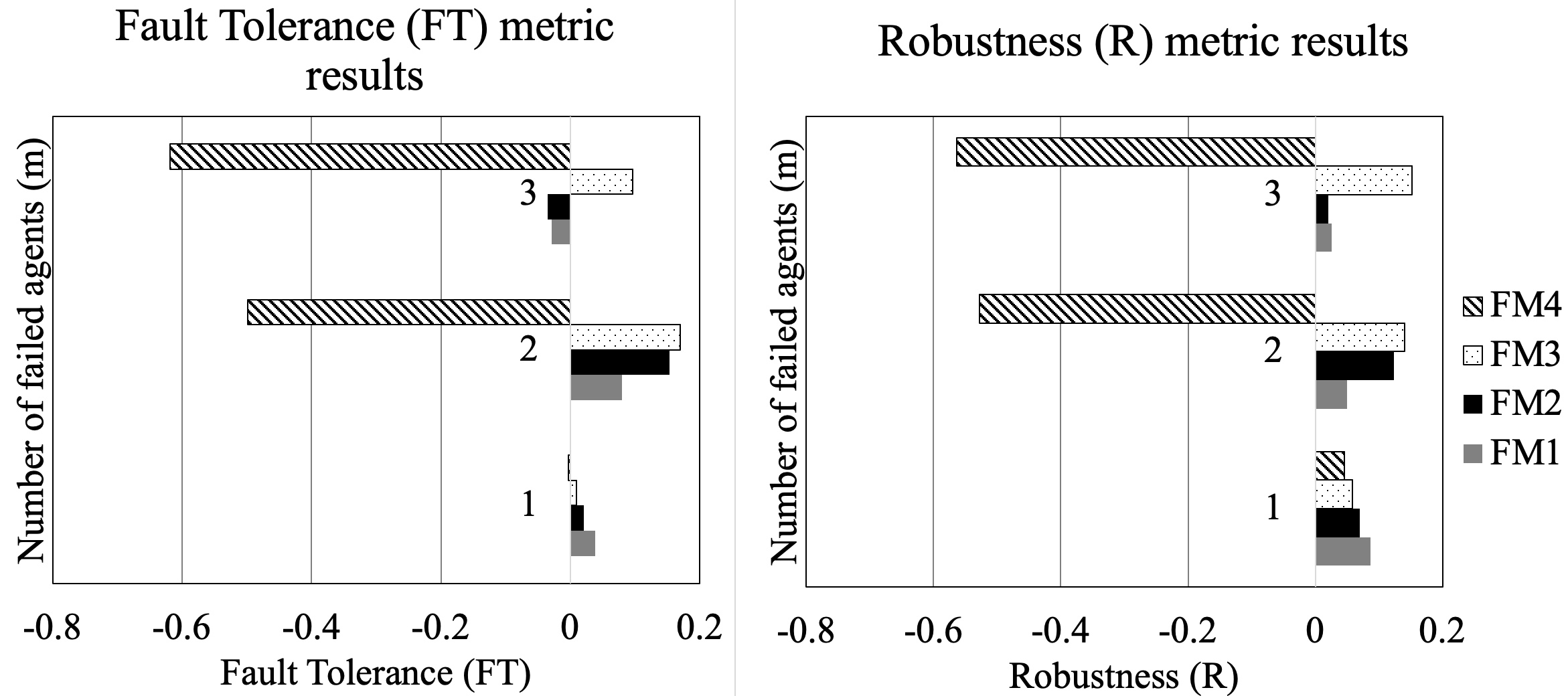}
\caption{Fault Tolerance (FT) and Robustness (R) metric results for 1-3 failed agents by failure modes 1-4.}
\label{fig:RFT}
\end{figure}

\begin{figure}[h]
   \centering
\includegraphics[scale=0.1250]{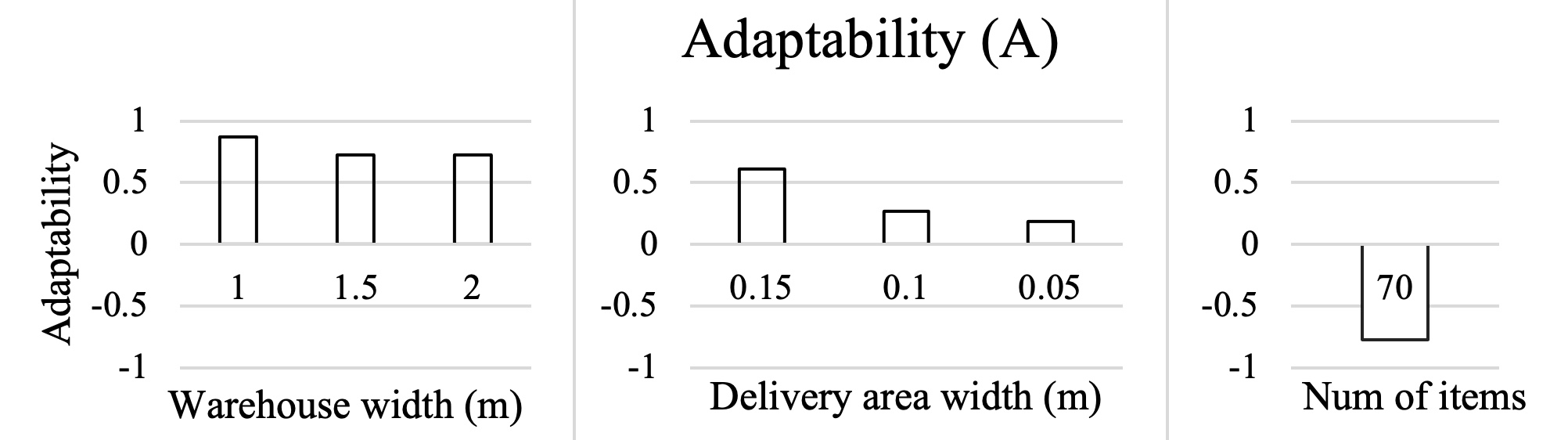}
\caption{Adaptability metric, A results for the changing parameters}
\label{fig:A}
\end{figure}

The Swarm Performance Indicators are here applied to an example use case to demonstrate their use in testing and their function in analysis of a system. The data used for the calculations was generated using a logistics scenario in a 2D physics-based simulator in Python. The robots are based on the Toshiba DOTs \cite{dots}. The arena includes a bounded 10 m by 10 m space which holds items in storage. The robot swarm is tasked with finding particular items (e.g. item ID $i_N$) for users who request them. The performance is measured by the number of items that are correctly delivered in a sequence of requests, in 500 s. The items are delivered when the robot carries them to a designated delivery area (the robots have no internal map of the warehouse). A swarm algorithm called Swarm Diffusion-Taxis (Swarm D-T) from \cite{swarmdiffusiontaxis} was used. In Swarm D-T the robots search the space with random walk and collision avoidance. When they are within range of an item they can identify its ID number and match it to the requested item ID. They periodically pick up and drop items at random with probabilistic sampling so as to reshuffle the space and avoid occlusions. With Swarm D-T, when a robot is in the delivery area, it is within range of a signal source. This robot will broadcast to its local neighbours a timer value of how recently it has been in the delivery area (timer = 0 s in the delivery area). This timer increases for every second it is out of the delivery area and is continually communicated with neighbours until it reaches 10 s when it stops and resets the timer to NaN. If a robot has the requested item then it will receive local neighbour timer values and move towards the robot with the lowest (none NaN) value. This creates a taxis effect towards the delivery area without the need for a global map or central control \cite{swarmdiffusiontaxis}.
\\
\textit{\textbf{Scalability}}\\
The researcher tests their swarm for several swarm sizes ($N$) and measures the performance ($P$) for each swarm size. These results are in Table \ref{tab:S} They then used Equation \ref{eq:SdP} to calculate the percentage change in performance followed by Equation \ref{eq:SdN} for the percentage change in swarm size from $N$ to $N+m$, for each increase in swarm size. Then the researcher uses Equation \ref{eq:S} to calculate the Scalability, S. The results for $S$ are given in Figure \ref{fig:S}.
%
\begin{table}[h]
\caption{Swarm sizes and their performance (items collected in 500s)}
\begin{tabular}{|l|l|l|l|l|l|l|l|l|l|}
\hline
\textbf{Swarm size (N)}  & 5 & 10 & 20 & 30 & 40 & 50 & 60 & 100 & 150 \\ \hline
\textbf{Performance (P)} & 9 & 18 & 29 & 33 & 45 & 56 & 65 & 74  & 69  \\ \hline
\end{tabular}
\label{tab:S}
\end{table}
Using the $S$ values in Figure \ref{fig:S}, a user of the swarm can read that the swarm is scalable ($S > 0$) for all gaps tested from 5 to 100 robots. They know that this means that they can keep expecting an increase in performance when they add agents up to 100 robots. They know that there will be a decrease in performance around 150 agents, this is the limit of the swarm's scalability ($S < 0$), in this task and set up. If the user values efficiency then they can look up the scalability values for these specific $N$ to $N+m$ ranges and decide if they should add agents to the swarm given their current swarm size. For example, if a user has $N$ = 30 agents then they can read that they will get a superscalable increase in performance if they increase to $N$ = 40 agents because here $S > 1$. This is more efficient than say, going from 10 to 20 agents, where the performance will increase but $0 < S < 1$ and will therefore not be a superlinear increase in performance.\\
\\
\textit{\textbf{Fault Tolerance and Robustness}}\\
To test Fault Tolerance and Robustness, the researcher knew that it was important to test multiple failure modes separately. They considered their swarm system and decided that the following were the most likely failure modes: 
\begin{itemize}
    \item Failure Mode 1 (FM1): Faulty agents always communicate that they are in the delivery area  area.
    \item Failure Mode 2 (FM2): Faulty agents cannot pick up items.
    \item Failure Mode 3 (FM3): Faulty agents cannot communicate with their neighbours. 
    \item Failure Mode 4 (FM4): Faulty agents cannot deliver items. This causes deadlocks because item $i_{N}$ cannot be delivered until $i_{N-1}$ has been delivered. 
\end{itemize}
\textbf{Fault tolerance} To calculate the Fault Tolerance to each failure mode, the researcher first measures the performance without any failures using $N$ = 25 robots. This is $P_o$, which is used later. Then the researcher simulates each failure mode, in separate experiments with 1 robot failing by each failure mode. The performance for Failure Mode X gives performance under faults, $P_F$. This is then used with $P_o$ in Equation \ref{eq:dPF} to get $\%\Delta P_F$. Finally, the researcher simulates the experiment again, this time with a swarm of 24 robots, as if the failed agents had been removed from the space before they could have an effect. This is used to calculate $\%\Delta P_S$ using Equation \ref{eq:dS}. The Fault Tolerance, FT, is then calculated using Equation \ref{eq:FT}. This was repeated for 2 and then 3 robot failures ($m$), giving new values for $P_F$ and $P_S$. The results used in these calculations are in Table \ref{tab:RFTdata} and the $FT$ values are in Figure \ref{fig:RFT}. It was found that the swarm was fault tolerant ($FT > 0$) to FM1 and FM2 up to 2 robot failures, up to 3 failures for FM3 and not fault tolerant to any failures by FM4.\\
\textbf{Robustness} to calculate the Robustness to each failure mode, Equation \ref{eq:RdN} is used to calculated the $\%\Delta N$ for $N=$ 25 agents with $m=$ 1, 2 and 3 agent failures. Each failure mode is then simulated to give $P_F$ which was combined with the $P_o$ (performance with no failures and 25 agents) to give $\%\Delta P$ using Equation \ref{eq:RdP} (these results are the same as were used for the $FT$ calculations and are also given in Table \ref{tab:RFTdata}). It was found that the swarm was robust ($R > 0$) to 3 agent failures for FM1, FM2 and FM3, and robust to 1 agent failure for FM4. The full results for $R$ are in Figure \ref{fig:RFT}. 
%
\begin{table}[h]
\caption{Number of failed agents, $m$, and performances, $P_F$, (items collected) for failure modes FM1-4. $P_S$ is performance with $N-m$ robots. Also given is percentage change in swarm size $\%\Delta N$.}
\begin{tabular}{|l|l|l|l|l|l|l|ll}
\cline{1-7}
\textbf{m} & \textbf{$\%\Delta N$} & \textbf{$P_S$} & \textbf{FM1 $P_F$} & \textbf{FM2 $P_F$} & \textbf{FM3 $P_F$} & \textbf{FM4 $P_F$} &  &  \\ \cline{1-7}
0          & -                 & 39          & 39              & 39              & 39              & 39              &  &  \\ \cline{1-7}
1          & 0.04               & 39          & 41               & 40              & 40              & 39               &  &  \\ \cline{1-7}
2          & 0.08               & 35          & 38              & 41              & 41              & 15              &  &  \\ \cline{1-7}
3          & 0.12              & 36          & 35             & 35              & 40               & 12              &  &  \\ \cline{1-7}
\end{tabular}
\label{tab:RFTdata}
\end{table}
\\
\textit{\textbf{Adaptability}}\\
The swarm ($N$ = 25) was tested for adaptability to three different conditions, where in each case one parameter $x$ was changed and the performance change as a result was measured. These were: (1) warehouse width; (2) delivery area width; (3) total number of items in storage. The resulting Adaptability, $A$, values calculated for each are given in Figure \ref{fig:A}. For the changing warehouse width the starting width, from which the change was measured for adaptability, was 0.5 m ($x_o$). The performance was measured with this width, giving $P_o$ then the width was increased to 1, 1.5 and 2 m ($x_{new}$) and the performances ($P_{new}$) measured again, with all other set up parameters (including delivery area  width and number of items) remaining the same. $A$ was calculated using Equation \ref{eq:A} with percentage change in $x$ from Equation \ref{eq:Adx} and the percentage change in performance from Equation \ref{eq:AdP}. 
\begin{table}[h]
\caption{Performance, $P$ (items collected in 500s) for each parameter value ($x$)}
\begin{tabular}{|l|l|l|l|l|lll}
\cline{1-2} \cline{4-5} \cline{7-8}
\textbf{\begin{tabular}[c]{@{}l@{}}Warehouse \\ width (x)\end{tabular}} & \textbf{P} &  & \textbf{\begin{tabular}[c]{@{}l@{}}Delivery\\ area \\ width (x)\end{tabular}} & \textbf{P} & \multicolumn{1}{l|}{} & \multicolumn{1}{l|}{\textbf{\begin{tabular}[c]{@{}l@{}}Number \\of \\items (x)\end{tabular}}} & \multicolumn{1}{l|}{\textbf{P}} \\ \cline{1-2} \cline{4-5} \cline{7-8} 
0.5 m ($x_o$)                                                                   & 47      &  & 0.2 m ($x_o$)                                                              & 41       & \multicolumn{1}{l|}{} & \multicolumn{1}{l|}{60 ($x_o$)}                                                                   & \multicolumn{1}{l|}{44}      \\ \cline{1-2} \cline{4-5} \cline{7-8} 
1 m                                                                     & 41       &  & 0.15 m                                                             & 37       & \multicolumn{1}{l|}{} & \multicolumn{1}{l|}{70}                                                                   & \multicolumn{1}{l|}{31}      \\ \cline{1-2} \cline{4-5} \cline{7-8} 
1.5 m                                                                   & 21      &  & 0.1 m                                                              & 26      &                       &                                                                                           &                                 \\ \cline{1-2} \cline{4-5}
2 m                                                                     & 8       &  & 0.05 m                                                             & 16      &                       &                                                                                           &                                 \\ \cline{1-2} \cline{4-5}
\end{tabular}
\label{tab:A}
\end{table}
Using the results in Figure \ref{fig:A}, the user can see that, with the test set up, the swarm is adaptable to a warehouse width from 0.5 ($x_o$) to 2 m. It is also adaptable, with another test set up, to an delivery area width from 0.2 ($x_o$) to 0.05 m. However it is not adaptable to changing number of items in the warehouse from 60 items ($x_o$).
\section{Discussion}
The Swarm Performance Indicators have been shown in the use case example to successfully decouple these traits from performance. For example, the logistics swarm set up is shown to not be adaptable for changing item number from 60 to 70 items. This does not mean the swarm is unusable as the performance goes from 44 to 31 items in 500 s, which could still be a reasonable performance for the user. The Adaptability metric describes how the performance change compares to the parameter change, and the result is that the performance changes more than the parameter change (i.e. not adaptable, $A < 0$). In addition, the SPIs provide important context and information alongside their metrics, beyond Boolean classifications which are reductionist without context (e.g. ``robust" or ``not robust"). The scalability metric includes a calculation using the percentage change in swarm size therefore it can only be given for a specific set up which should be quoted in full alongside the metric, to justify the scalability of a system. The fault tolerance and robustness metrics require a full set of experiments for each failure mode and are only relevant to each one. $FT$ also requires the scaled down swarm performance to be tested, which shows the effect of failed agents on their neighbours (by comparing this performance to the failed robot performance). The adaptability of the use case was tested for three different environmental parameters and could only be given when testing on these parameters is done, therefore cannot be used as a general metric to say e.g. ``the swarm is adaptable". This is a good thing as it requires context to be given for the metric, such as ``Swarm S is adaptable to changing warehouse width 0.5 to 2 m, for Task T in Warehouse W with set up parameter set C". This context also means that the metrics are generalisable to any testable swarm and encourages researchers to give details and experimental data to prove their system is robust, fault tolerant, scalable or adaptable.\\
\textbf{Limits} The Adaptability metric can only test the change in one given parameter, $x$ which means it cannot be used to test adaptability across different tasks or environments. There are definitions of adaptability in the literature (Section \ref{sec:lit_review}) that would dispute this as necessary for this definition, which may cause confusion. 
Finally, one feature of swarm fault tolerance and robustness that is not discussed here but is in the literature \cite{winfield_bjerknes} is their ability to self-repair their behaviour in the presence of faults. 
In the SPIs $FT$ and $R$, only the end performance result is used in the calculations and there is no discussion of the performance over time which could describe self-repair time. \\
\textbf{Conclusions} The Swarm Performance Indicators have successfully derived the otherwise qualitative traits of Robustness, Fault Tolerance, Scalability and Adaptability, into quantitative measures. These have been demonstrated on simulated results from a logistics use case that uses a robot swarm. 
\addtolength{\textheight}{-12cm} 
%
%
%
%
\bibliographystyle{siam}
\bibliography{bib}
\end{document}